\documentclass[11pt,a4paper]{article}
\usepackage[hyperref]{emnlp-ijcnlp-2019}
\usepackage{times}
\usepackage{latexsym}

\usepackage{url}

\usepackage{graphicx}
\usepackage{amsmath}
\usepackage{amsfonts}
\usepackage{bm} 
\usepackage{threeparttable}
\usepackage{booktabs}
\usepackage{algorithm} 
\usepackage{algpseudocode} 
\usepackage{CJKutf8}
\usepackage{multirow}
\usepackage{xcolor}
\usepackage{array}

\providecommand{\model}{SS}
\providecommand{\problem}{CSVD}
\providecommand{\emb}{VFGE}

\makeatother

\aclfinalcopy 

\title{Detect Camouflaged Spam Content via StoneSkipping: Graph and Text Joint Embedding for Chinese Character Variation Representation}

\author{Zhuoren Jiang$^1$\thanks{~~These two authors contributed equally to this research.}~, Zhe Gao$^2$\footnotemark[1]~, Guoxiu He$^3$, Yangyang Kang$^2$, Changlong Sun$^2$,\\ \textbf{Qiong Zhang$^2$,~Luo Si$^2$,~Xiaozhong Liu$^4$\thanks{~~Corresponding author}} \\
$^1$ School of Data and Computer Science, Sun Yat-sen University, Guangzhou, China\\
$^2$Alibaba Group, Hangzhou \& Sunnyvale \& Seattle, China \& USA\\
$^3$School of Information Management, Wuhan University, Wuhan, China\\
$^4$School of Informatics, Computing and Engineering, Indiana University Bloomington, Bloomington, USA\\
{\tt jiangzhr3@mail.sysu.edu.cn,}\\
{\tt\{gaozhe.gz,yangyang.kangyy,qz.zhang,luo.si\}@alibaba-inc.com,}\\
{\tt guoxiu.he@whu.edu.cn, changlong.scl@taobao.com, liu237@indiana.edu}}

\date{}

\begin{document}
\begin{CJK}{UTF8}{gbsn}
\maketitle
\begin{abstract}
 The task of Chinese text spam detection is very challenging due to both glyph and phonetic variations of Chinese characters. This paper proposes a novel framework to jointly model Chinese variational, semantic, and contextualized representations for Chinese text spam detection task. In particular, a Variation Family-enhanced Graph Embedding (VFGE) algorithm is designed based on a Chinese character variation graph. The VFGE can learn both the graph embeddings of the Chinese characters (local) and the latent variation families (global). Furthermore, an enhanced bidirectional language model, with a combination gate function and an aggregation learning function, is proposed to integrate the graph and text information while capturing the sequential information. Extensive experiments have been conducted on both SMS and review datasets, to show the proposed method outperforms a series of state-of-the-art models for Chinese spam detection.
\end{abstract}

\section{Introduction}
\begin{figure}[btp]\centering
 	\includegraphics[width=1.0\columnwidth]{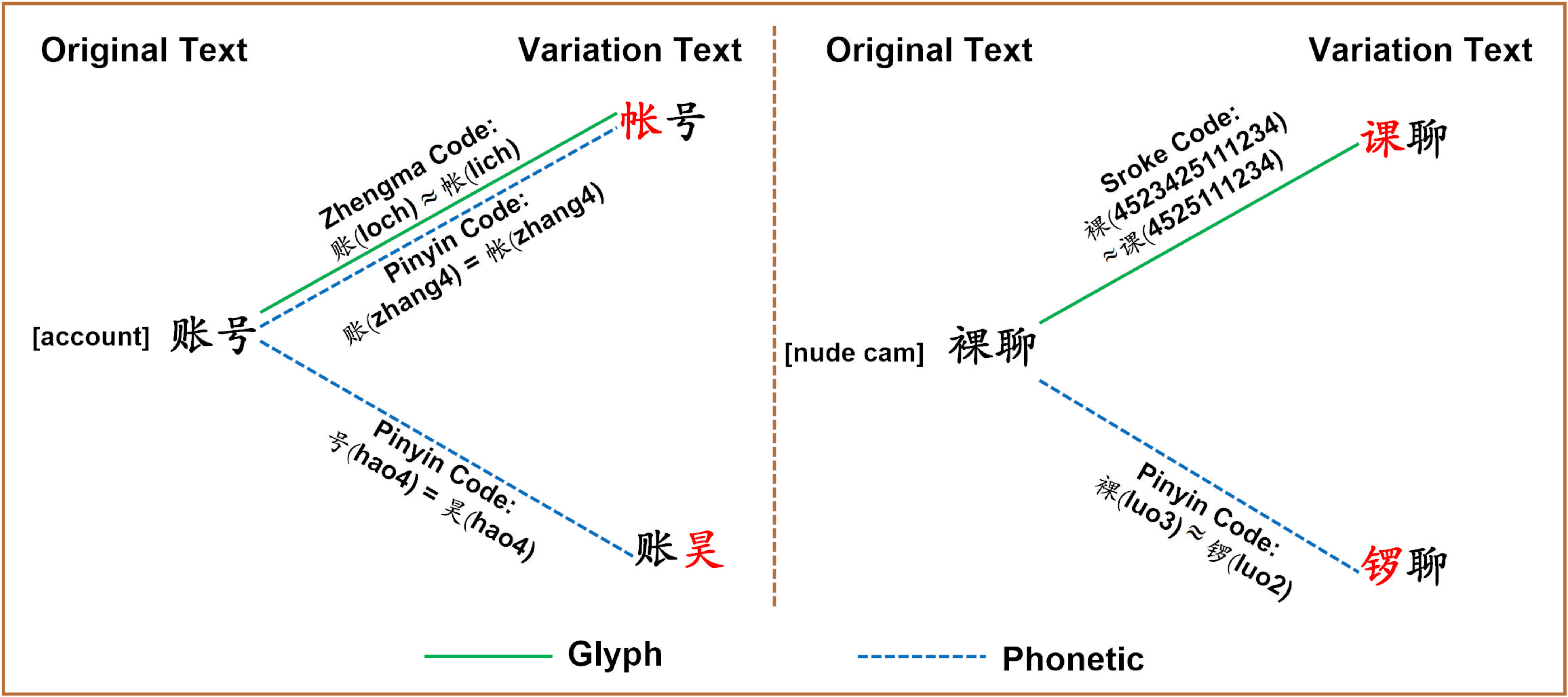}
 	\caption{Character Variations in Chinese Spam Texts (the pinyin codes provide phonetic information and the stroke and Zhengma codes provide glyph information).}
 	\label{fig:exp}
\end{figure}

Chinese orchestrates over tens of thousands of characters by utilizing their morphological information, e.g., pictograms, simple/compound ideograms, and phono-semantic compounds \cite{norman1988chinese}. Different characters, however, may share the similar glyph and phonetic ``root''. For instance, from glyph perspective, character ``裸 (naked)'' looks like ``课 (course)'' (homographs), while from phonetic viewpoint, it shares the similar pronunciation with ``锣 (gong)'' (homophones). The form of variations can also be compounded, for instance, ``账 (account)''and ``帐 (curtain)'' have the similar structure and pronunciation (homonyms). Unfortunately, in the context of spam detection, as shown in Figure \ref{fig:exp}, spammers are able to take advantage of these variations to escape from the detection algorithms \cite{jindal2007review}. For instance, in the e-commerce ecosystem, variation-based Chinese spam mutations thrive to spread illegal, misleading, and harmful information\footnote{More detailed information can be found in the experiment section.}. In this study, we propose a novel problem - Chinese Spam Variation Detection (\problem), a.k.a. investigating an effective Chinese character embedding model to assist the classification models to detect the variations of Chinese spam text, which needs to address the following key challenges.

\textbf{Diversity}: the variation patterns of Chinese characters can be complex and subtle, which are difficult to generalize and detect. For instance, in the experimental dataset, one Chinese character can have 297 (glyph and phonetic) variants averagely and 2,332 maximally. The existing keyword based spam detection approaches, e.g., \cite{ntoulas2006detecting}, can hardly address this problem. \textbf{Sparseness, Zero-shot, and Dynamics}: when competing with classification models, spammers are constantly creating new Chinese characters combinations for spam texts (that can be a ``zero/few shot learning'' problem \cite{socher2013zero}). The labelling cost can be inevitably high in such dynamic circumstance. Data driven approaches, e.g., \cite{zhang2015character}, will perform poorly for unseen data. \textbf{Camouflage}: with the common cognition knowledge of Chinese and the contextual information, users are able to consume the spam information, even when some characters in the content are intentionally mutated into their similar variations \cite{spinks2000reading,shu1997role}. However, the variation-based spam text are highly camouflaged for machines. It is important to propose a novel Chinese character representation learning model that can synthesize character variation knowledge, semantics, and contextualized information.

To address these challenges, we propose a novel solution, \textbf{S}tone\textbf{S}kipping (\textbf{\model}) model to enable Chinese variation representation learning via graph and text joint embedding. {\model} is able to learn the Chinese character variation knowledge and predict the new variations not appearing in the training set by utilizing sophisticated heterogeneous graph mining method. For a piece of text (a character sequence), with the proposed model, each candidate character can probe character variation graph (like stone bouncing cross the water surface), and explore its glyph and phonetic variation information (like the ripples caused by the stone hitting the water).  Algorithmically, a Variation Family-enhanced Graph Embedding (\emb) algorithm is proposed to extract the heterogeneous Chinese variation knowledge while learning the (local) graph representation of a Chinese character along with the (global) representation of the latent variation families. Finally, an enhanced bidirectional language model, with a combination gate function and an aggregation learning function, is proposed to comprehensively learn the variation, semantic, and sequential information of Chinese characters. To the best of our knowledge, this is the first work to use graph embedding to learn the heterogeneous variation knowledge of Chinese characters for spam detection.

The major contributions of this paper can be summarized as follows: 

1. We propose an innovative {\problem} problem, in the context of text spam detection, to address the diversity, sparseness, and text camouflage problems.

2. A novel joint embedding {\model} model is proposed to learn the variational, semantic, and contextual representations of Chinese characters. {\model} is able to predict unseen variations.

3. A Chinese character variation graph is constructed for encapsulating the glyph and phonetic relationships among Chinese characters. Since the graph can be potentially useful for other NLP tasks, we share the graph/embeddings to motivate further investigation. 

4. Through the extensive experiments on both SMS and review datasets\footnote{In order to help other scholars reproduce the experiment outcome, we will release the datasets via GitHub (https://github.com/Giruvegan/stoneskipping)}, we demonstrate the efficacy of the proposed method for Chinese spam detection. The proposed method outperforms the state-of-the-art models.

\section{Related Work}\label{review}
\textbf{Neural Word Embeddings}. Unlike traditional word representations, low-dimensional distributed word representations~\cite{mikolov2013distributed,pennington2014glove} are able to capture in-depth semantics of text content. More recently, ELMo~\cite{peters2018deep} employed learning functions of the internal states of a deep bidirectional language model to generate the character embeddings. BERT~\cite{devlin2018bert} utilized bidirectional encoder representations from transformers~\cite{vaswani2017attention} and achieved improvements for multiple NLP tasks. However, all the prior models only focused on learning the context, whereas the text variation was ignored. Moreover, {\problem} problem can be different from other NLP tasks: the intentional character mutations and unseen variations (zero-shot learning \cite{socher2013zero}) can threaten prior models' performances.

\textbf{Chinese Word and Sub-word Embeddings}. A number of studies explored Chinese representation learning methodologies. CWE~\cite{chen2015joint} learned the character and word embeddings to improve the representation performance. GWE~\cite{su2017learning} introduced the features extracted from the images of traditional Chinese characters. JWE~\cite{yu2017joint} used deep learning to generate character embedding based on an extended radical collection. Cw2vec~\cite{cao2018cw2vec} investigated Chinese character as a sequence of n-gram stroke order to generate its embedding. Although these models had considered the nature of Chinese characters, they only utilized glyph features while the phonetic information was ignored. In {\problem} problem, the forms of variations can be heterogeneous, and a single kind of features cannot cover all mutation patterns. More importantly, all these models are not designed for spam detection, and the task-oriented model should be able to highlight the most important variations for spam text.  

\textbf{Graph Embedding}. Graph (a.k.a. information network) is a natural data structure for characterizing the multiple relationships between the objects. Recently, multiple graph embedding algorithms are proposed to learn the low dimensional feature representations of vertexes in graphs. DeepWalk~\cite{perozzi2014deepwalk} and Node2vec~\cite{grover2016node2vec} are random walk based models. LINE \cite{tang2015line} modeled 1st and 2nd order graph neighbourhood. Meanwhile, metapath2vec++~\cite{dong2017metapath2vec} was designed for heterogeneous graph embedding with human defined metapath rules. HEER \cite{shi2018easing} is a recent state-of-the-art heterogeneous graph embedding model. Though the techniques utilized in these models are different, most existing graph embedding models focus more on local graph structure representation, e.g., modelling of a fixed-size graph neighbourhood. {\problem} problem requires graph embedding conducted from a more global perspective, to characterize comprehensive variation patterns.

\textbf{Spelling Correction}. Spelling correction may serve as an alternative to address {\problem} problem, e.g., using dictionary-based \cite{yeh2014chinese} or language model-based method \cite{yu2014chinese} to restore the content variations to their regular format. However, because spammers intentionally mutate the spam text to escape from the detection model, training data sparseness and dynamics may challenge this approach.

\begin{figure*}[]\centering
 	\includegraphics[width=2.0\columnwidth]{pics/method_new_new_new.pdf}
 	\caption{An Illustration of ``StoneSkipping'' Framework}
 	\label{fig:frame}
\end{figure*}

\section{StoneSkipping Model}\label{sec:method}
Figure \ref{fig:frame} depicts the proposed {\model} model. There are three core modules in {\model}: a \textbf{Chinese character variation graph} to host the heterogeneous variation information; a \textbf{variation family-enhanced graph embedding} for Chinese character variation knowledge extraction and graph representation learning; an \textbf{enhanced bidirectional language model} for joint representation learning. In the remaining of this section, we will introduce them in detail.

\subsection{Chinese Character Variation Graph}
A Chinese character variation graph\footnote{We will release Chinese character variation graph via GitHub (https://github.com/Giruvegan/stoneskipping).} can be denoted as $G=(C,R)$. $C$ denotes the Chinese character set, and each character is represented as a vertex in $G$. $R$ denotes the variation relation (edge) set, and edge weight is the similarity of two characters given the target relation (variation) type. To accurately characterize both phonetic and glyph information of Chinese character, we utilize three different encoding methods: 

\textbf{Pinyin} system provides phonetic-based information, which is widely used for representing the pronunciations of Chinese characters~\cite{chen2000new}. In this system, each Chinese character has one syllable which consists of three components: an initial (consonant), a final (vowel), and a tone. There are four types of tones in Modern Standard Mandarin Chinese. Different tones with the same syllable can have different meanings. For instance, the pinyin code of ``裸 (naked)'' is ``luo3'' and ``锣 (gong)'' is `luo2''. The pinyin-based variation similarity is calculated based on their pinyin syllables with tones\footnote{Because of the space limitation, the detailed operations of relation generation will be provided on https://github.com/Giruvegan/stoneskipping.\label{dg}}.  

\textbf{Stroke} is a basic glyph pattern for writing Chinese character~\cite{cao2018cw2vec}. All Chinese characters are written in a certain stroke order and can be represented as a stroke code, e.g., the stroke code of ``裸 (naked)'' is ``4523425111234'' and `课 (course)'' is ``4525111234''. The stroke-based variational similarity is calculated based on \textit{longest common substring} and \textit{longest common subsequence} metrics\textsuperscript{\ref {dg}}.

\textbf{Zhengma} is another important means for glyph character encoding, which encodes character at radical level~\cite{yu2017joint}. For instance, the Zhengma code of ``裸 (naked)'' is ``WTKF'' and `课 (course)'' is ``SKF''. The Zhengma-based variational similarity is calculated based on the \textit{Jaccard Index} metric\textsuperscript{\ref{dg}}.

Unlike previous works \cite{cao2018cw2vec,yu2017joint} only employ one kind of glyph-based information, we utilize two different glyph patterns (stroke and Zhengma) to encode the Chinese character. Because these two patterns can characterize Chinese characters from different internal structural levels, and complement each other to enable an enhanced glyph representation learning. Furthermore, the pinyin encoder provides phonetic information. The constructed character variation graph integrates these three kinds of variation relations, which can be significant for camouflaged spam detection.

\subsection{Variation Family-enhanced Graph Embedding}
While the variation graph can provide comprehensive knowledge of Chinese character variations, efforts need to be made to address these two problems: (1) the variation patterns can be very flexible, and the compounded (long-range) variation information transfer may exist. Therefore, short-range (local) graph information, e.g., character vertex's neighbors, may be insufficient for spam detection. Meanwhile, it is impractical to exhaust all the possible variation patterns. (2) To oblige users to consume the text content, spammers cannot make the variation patterns to be too complex/confusing, they usually focus on the most sensitive words in a spam message. Hence, some random infrequent variation patterns could be ``noisy'' for {\problem} while polluting the detection outcomes.

\textbf{Latent Character Variation Family}. In this study, we propose a {\emb} model to address these problems. As depicted in Figure \ref{fig:frame}, in {\emb} model, we introduce a set of latent variables ``\textit{character variation family}'' $F = \left \{ F_{1},...,F_{|F|} \right \}$ at a graph schema (global) level to capture the critical information for spam detection. Each $F_{i}$ is defined as \textit{a distribution of characters}, which aims to estimate the global frequent variation dependencies in $G$. By learning $F$, {\emb} is able to highlight the useful variations, eliminate the noisy patterns, and predict the unseen variation forms w.r.t. the spam detection task. 

\textbf{Random Walk based Character Family Representation Co-Learning }. {\emb} is a random walk based graph embedding model, and we employ a hierarchical random walk strategy \cite{jiang2018cross} on $G$ to generate the optimized walking paths (character vertex sequences) for each character. The model can sample the most possible variation context vertexes for each character. Based on generated walking paths, {\emb} executes the following two processes iteratively:

(1) \textbf{Family Assignment}. By leveraging both local context and global family distributions, we assign a discrete family for each character vertex in a particular walking path to form a character-family pair $\left \langle C,F \right \rangle$. As shown in Figure \ref{fig:frame}, we assume different walking paths tend to emerge various character variation patterns which can be represented as mixtures over latent variation families. Given a character $C_{i}$ in a $path$, $C_{i}$ has a higher chance to be assigned to a dominant family $F_{i}$. The assignment probability can be calculated as:
\begin{equation}
\begin{aligned}
& Pr(F_{i}|C_{i},path) \\  \propto & Pr(C_{i},F_{i},path) \\ = & Pr(path) Pr(F_{i}|path) Pr(C_{i}|F_{i})
\end{aligned}
\end{equation}
As depicted in Figure \ref{fig:frame}, $\alpha$ is the parameter of the Dirichlet prior on the per-path family distributions ($Pr(path)$); $\beta$ is the family assignment distribution ($Pr(C|F)$); and $\theta$ is the family mixture distribution for a walking path ($Pr(F|path)$). The distribution learning can be considered as a Bayesian inference problem, and we use Gibbs sampling \cite{porteous2008fast} to address this problem. 

(2) \textbf{Character-Family Representation Co-Learning}. Given the assigned character-family pairs, the proposed method aims to obtain the representations of character $C$ and latent variation family $F$ by mapping them into a low-dimensional space $\mathbb{R}^{d}$ ($d$ is a parameter specifying the number of dimensions). Motivated by \cite{liu2015topical}, we propose a novel representation learning method to optimize characters and families \textbf{\textit{separately}} and \textbf{\textit{simultaneously}}.

The objective is defined to maximize the following log probability:
\begin{equation}\label{equ:max}
\mathcal{L} = \underset{f}{max}\sum_{C_{i}\in C} \sum_{C_{j} \in \mathbb{N}(C_{i})} log Pr(\left \langle C_{j},F_{j} \right \rangle | \mathbf{C_{i}^{F_{i}}})
\end{equation}
We use $f(\cdot)$ as the embedding function. $\mathbf{C_{i}} = f(C_{i})$ represents the character graph embedding and $\mathbf{F_{i}} = f(F_{i})$ represents the family graph embedding.  $\mathbf{C_{i}^{F_{i}}}$ denotes the concatenation of $\mathbf{C_{i}}$ and $\mathbf{F_{i}}$, whereas $\mathbb{N}(C_{i})$ is $C_{i}$'s neighborhood (context). As Figure \ref{fig:frame} shows, the feature representation learning method is an upgraded version of the skip-gram architecture. Compared with merely using the target vertex $C_{i}$ to predict context vertexes in original skip-gram model~\cite{mikolov2013distributed}, the proposed approach employs the character-family pair $\left \langle C_{i},F_{i} \right \rangle$ to predict context character-family pairs. From variation viewpoint, character vertex's context will encapsulate both local (vertex) and global (variation family) information. Hence, the learned representations are able to comprehensively preserve the variation information in $G$.

$Pr(\left \langle C_{j},F_{j} \right \rangle | \mathbf{C_{i}^{F_{i}}})$ is modeled as a softmax function:
\begin{equation}\label{equ:softmax1}
Pr(\left \langle C_{j},F_{j} \right \rangle | \mathbf{C_{i}^{F_{i}}})= \frac{exp(\mathbf{C_{j}^{F_{j}}} \cdot \mathbf{C_{i}^{F_{i}}})}{\sum_{C_{k} \in C}exp(\mathbf{C_{k}^{F_{k}}} \cdot \mathbf{C_{i}^{F_{i}}})}
\end{equation}
Stochastic gradient ascent is used for optimizing the model parameters of $f$. Negative sampling \cite{mikolov2013distributed} is applied for optimization efficiency. Note that, the parameters of each character embedding and family embedding are shared over all the character-family pairs, which, as suggested in~\cite{liu2015topical}, can address the training data sparseness problem and improve the representation quality.

\textbf{Family-enhanced Embedding Integration}. As shown in Figure \ref{fig:frame}, the family-enhanced character graph embedding can be calculated as: 
\begin{equation}
\mathbf{G_{i}}=  \left [\mathbf{C_{i}}, \sum _{F_{j} \in F} Pr(F_{j}|C_{i})\mathbf{F_{j}}\right ]
\end{equation}
where $\mathbf{G_{i}}$ is family-enhanced graph embedding for $C_{i}$, and $\left [ \cdot \right ]$  is concatenating operation. $Pr(F_{j}|C_{i})$ can be inferred from family assignment distribution $\beta$.

\begin{table*}[htbp]
\small
\centering
\begin{tabular}{cp{0.35\textwidth}ccp{0.05\textwidth}cc}
\toprule
\multirow{2}*{\textbf{Group}}  & \multirow{2}*{\textbf{Model}} &
\multicolumn{2}{c}{\textbf{SMS}} & & \multicolumn{2}{c}{\textbf{Review}}\\ 
&&\textbf{Accuracy} & \textbf{F1 Score} & &\textbf{Accuracy} & \textbf{F1 Score}\\
\midrule
\multirow{3}*{\textbf{Text}}
&\textbf{Skipgram}~\cite{mikolov2013distributed} & 0.807 & 0.765 & & 0.693  & 0.560   \\
&\textbf{GloVe}~\cite{pennington2014glove} & 0.732 & 0.637 & & 0.707 & 0.600 \\
&\textbf{ELMo}~\cite{peters2018deep}& 0.786 & 0.747 & &  0.755 & 0.647 \\
\midrule
\multirow{4}*{\textbf{Chinese}}
&\textbf{CWE}~\cite{chen2015joint} & 0.751 & 0.674 & & 0.780 & 0.726 \\
&\textbf{GWE}~\cite{su2017learning} & 0.505 & 0.426 & & 0.778 & 0.718\\
&\textbf{JWE}~\cite{yu2017joint} & 0.770 & 0.707 & & 0.738 & 0.646\\
&\textbf{Cw2vec}~\cite{cao2018cw2vec} & 0.800 & 0.753 & & 0.724 & 0.618  \\
\midrule
\multirow{5}*{\textbf{Graph}}
&\textbf{DeepWalk}~\cite{perozzi2014deepwalk} & 0.836 & 0.804 & & 0.738 & 0.638 \\
&\textbf{LINE}~\cite{tang2015line} & 0.821 & 0.783 & & 0.764 & 0.695\\
&\textbf{Node2vec}~\cite{grover2016node2vec} & 0.835 & 0.802 & & 0.792 & 0.736\\
&\textbf{M2V$\mathbf{_{Max}}$}~\cite{dong2017metapath2vec} &  0.838 & 0.807 & & 0.790 & 0.740 \\
&\textbf{HEER}~\cite{shi2018easing} & 0.723 & 0.617 & & 0.771 & 0.708\\
\midrule
\textbf{Correction}&\textbf{Pycorrector}~\cite{yu2014chinese}& 0.782  & 0.727 &  & 0.688 & 0.549 \\
\midrule
\multirow{3}*{\textbf{Comparison}}
&\textbf{\model}$\mathbf{_{Graph}}$ & 0.839 & 0.827 & & 0.812 & 0.756\\
&\textbf{\model}$\mathbf{_{Naive}}$ & 0.849 & 0.825 & & 0.811 & 0.757 \\
&\textbf{\model}$\mathbf{_{Original}}$ & \textbf{0.851} & \textbf{0.832} & & \textbf{0.854}  & \textbf{0.822}\\
\bottomrule
\end{tabular}
\caption{Chinese Text Spam Detection Performance Comparison of Different Models}
\label{tab:result}
\end{table*}

\subsection{Enhanced Bidirectional Language Model}
As shown in Figure \ref{fig:frame}, {\model} model utilizes an enhanced bidirectional language model to jointly learn variation, semantic and contextualized representation of Chinese character.

\textbf{Combination Gate Function}. This gate function is utilized for combining the variation and semantic representations, which is the input function for bidirectional language model. The formulations of the gate function are listed as follows:
\begin{equation}
\begin{aligned}
\mathbf{P} & = \sigma (\mathbf{W}_{P} \cdot \left [\mathbf{G}, \mathbf{T} \right ]+\mathbf{b}_{P})\\
\mathbf{N} & = (\mathbf{P} \odot \mathbf{T} ) + ((\mathbf{1}-\mathbf{P}) \odot \mathbf{G})
\end{aligned}
\end{equation}
$\mathbf{P}\in \mathbb{R}^{d}$ is the preference weights for controlling the contributions from $\mathbf{G} \in \mathbb{R}^{d}$ (variation graph embedding) and $\mathbf{T} \in \mathbb{R}^{d}$ (Skip-Gram textual embedding). $\mathbf{W}_{P} \in \mathbb{R}^{2d \times d}$. $\mathbf{N} \in \mathbb{R}^{d}$ is the combination representation. $\odot$ is elementwise product, and $+$ is elementwise sum.
 
\textbf{Aggregation Learning Function}. With the combination representation $\mathbf{N}$ as input, we train a bidirectional language model for capturing the sequential information. There could be multiple layers of forward and backward LSTMs in bidirectional language model. For $k_{th}$ character, $\overrightarrow{\mathbf{H}_{l}^{k}}$ is the forward LSTM unit's output for layer $l$, where $l = 1, 2, ..., L$, and $\overleftarrow{\mathbf{H}_{l}^{k}}$ is the output of the backward LSTM unit.

The output $\mathbf{SS}$ embedding is learned from an  aggregation function, which aims to aggregate the intermediate layer representations of the bidirectional language model and the input embedding $\mathbf{N}$. For $k_{th}$ character, if we denote $\mathbf{H}_{0}^{k} = \lbrack \mathbf{N}^{k}, \mathbf{N}^{k} \rbrack$ (self concatenation), and $\mathbf{H}_l^{k} = \lbrack \overleftarrow{\mathbf{H}_l^{k}}, \overrightarrow{\mathbf{H}_l^{k}} \rbrack$, the output can be:
\begin{equation}
\mathbf{SS^{k}} = \omega \big(\underbrace{s_0 \mathbf{H}_{0}^{k}}_{(\textup{Variational}~\&~ \textup{Semantic})}+~~~~\underbrace{\sum_{l=1}^{L}s_l \mathbf{H}_l^{k} }_{\textup{Contextualized}}\big)
\end{equation}
where $\omega$ is the scale parameter, and $s_l$ is a weight parameter for the combination of each layer, which can be learned through the training process. Similar aggregation operation has been proven useful to model the contextualized word representation \cite{peters2018deep}.




\section{Experiment}
\subsection{Dataset and Experiment Setting}\label{ssec:data}
\textbf{Dataset}\footnote{https://github.com/Giruvegan/stoneskipping}. In Table \ref{tab:data}, we summarize the statistics of the two real-world spam datasets (in Chinese). One is a SMS dataset, the other is a product review dataset. Both datasets were manually labeled (spam or regular labels) by professionals. False advertising and scam information are the most common forms of spam information for SMS dataset, while abuse information dominates review spam dataset.
\begin{table}[h]
\small
\centering
\begin{tabular}{ccccc}  
\toprule
\textbf{Dataset} & \textbf{Part} & \textbf{All} & \textbf{Spam} & \textbf{Normal} \\
\midrule
\multirow{2}*{SMS} & Train  & 48,884 & 23,891 & 24,993      \\
                   & Test & 48,896 & 23,891 & 25,005 \\
\midrule
\multirow{2}*{Review} & Train & 37,299 & 17,299 & 20,000    \\
                   & Test & 37,299 & 17,299 & 20,000 \\                   
\bottomrule
\end{tabular}
\caption{Statistics of Two Chinese Spam Text Datasets}
\label{tab:data}
\end{table}

In the constructed variation graph, there are totally 25,949 Chinese characters (vertexes) and 7,705,051 variation relations. For all the variation relations, there are 1,508,768 pinyin relations (phonetic), 373,803 stroke relations (glyph), and 5,822,480 Zhengma relations (glyph).

\begin{table*}[htbp]
\small
\centering
\begin{threeparttable}
\begin{tabular}{cllllllll}  
\toprule
\multirow{2}*{\textbf{Character}} &\multicolumn{2}{c}{\textbf{Text}} & \multicolumn{2}{c}{\textbf{Chinese}} &  \multicolumn{2}{c}{\textbf{Graph}} & \multicolumn{2}{c}{\textbf{Proposed model}}  \\ &\multicolumn{2}{c}{\textbf{Skipgram}} & \multicolumn{2}{c}{\textbf{Cw2vec}} &  \multicolumn{2}{c}{\textbf{\emb}} & \multicolumn{2}{c}{\textbf{\model}}  \\
\midrule
\multirow{3}*{运(move)} &
\colorbox{yellow}{C}&捷(prompt)
& \colorbox{yellow}{C}&捷(prompt)
&\colorbox{green}{G}\colorbox{cyan}{P}& 云(cloud)
&\colorbox{orange}{S}\colorbox{yellow}{C}&转(transmit)\\
&\colorbox{yellow}{C}&站(stop)
&\colorbox{yellow}{C}&站(stop)
&\colorbox{green}{G}\colorbox{cyan}{P}&纭(numerous)
&\colorbox{green}{G}\colorbox{cyan}{P}&芸(weed) \\
&\colorbox{yellow}{C}&客(guest)
&\colorbox{orange}{S}\colorbox{yellow}{C}&输(transport)
&\colorbox{green}{G}&坛(altar)
&\colorbox{green}{G}\colorbox{cyan}{P}&云(cloud) \\
\midrule
\multirow{3}*{惊(shock)} 
&\colorbox{orange}{S}\colorbox{yellow}{C}&讶(surprised)  &\colorbox{orange}{S}\colorbox{yellow}{C}&讶(surprised)
&\colorbox{green}{G}\colorbox{cyan}{P}&景(view)
&\colorbox{orange}{S}\colorbox{yellow}{C}&慌(flurried)\\
&\colorbox{orange}{S}\colorbox{yellow}{C}&愕(startled)
&\colorbox{orange}{S}\colorbox{yellow}{C}&撼(shake)
&\colorbox{green}{G}&晾(dry)
&\colorbox{green}{G}&琼(jade)\\ 
&\colorbox{orange}{S}\colorbox{yellow}{C}&吓(scare)
&\colorbox{orange}{S}\colorbox{yellow}{C}&愕(startled)
&\colorbox{green}{G}&谅(forgive)
&\colorbox{green}{G}\colorbox{orange}{S}\colorbox{yellow}{C}&悚(afraid)\\ 
\bottomrule
\end{tabular}
\begin{tablenotes}
    \footnotesize
    \item \colorbox{green}{G}: Glyph; \colorbox{cyan}{P}: Phonetic; \colorbox{orange}{S}: Semantic; \colorbox{yellow}{C}:Context
    \end{tablenotes}
\end{threeparttable}
\caption{Case Study: given the target character, we list the top 3 similar characters from each algorithm. The characters are selected from a frequently used candidate character set whose size is 8238.}
\label{tab:case}
\end{table*}

\textbf{Experimental Set-up}. We validated the proposed model in Chinese text spam detection task. In order to simulate the ``diversity'', ``sparseness'' and `` zero-shot'' problems under real business scenarios, we made a challenging restriction on the training and testing sets, i.e., the character variations were only included in testing set, and all samples in training set were using the original characters.

For the proposed {\model} model, we utilized the following setting: layers of LSTMs: 2; dimension of hidden (output) state in LSTM: 128; dimension of pre-trained character text embedding: 128; dimension of {\emb} embedding: 128; batch size: 64; Dropout: 0.1. For training {\emb} embedding\footnote{For the experiment fairness, all the random walk based graph embedding baselines shared the same parameters with {\emb}.}, the walk length was 80, the number of walks per vertex was 10. These parameters were adopted in \cite{peters2018deep,jiang2018mathematics,perozzi2014deepwalk,grover2016node2vec}. The variation family number\footnote{Based on the parameter sensitive analysis, the proposed method was not very sensitive to number of variation families.} was 500. {\model} model was pre-trained for parameter initialization as suggested in \cite{peters2018deep}.


\textbf{Baselines and Comparison Groups}. We chose 13 strong baseline algorithms, from text or graph viewpoints, to comprehensively evaluate the performance of the proposed method. 

\textit{General Textual Based Baselines}: \textbf{Skipgram}~\cite{mikolov2013distributed}, \textbf{GloVe}~\cite{pennington2014glove}, and \textbf{ELMo}~\cite{peters2018deep}.

\textit{Chinese Specific Textual Based Baselines}: \textbf{CWE}~\cite{chen2015joint}, \textbf{GWE}~\cite{su2017learning}, \textbf{JWE}~\cite{yu2017joint}, and  \textbf{Cw2vec}~\cite{cao2018cw2vec}.

\textit{Graph Embedding Based Baselines}: 
\textbf{DeepWalk}~\cite{perozzi2014deepwalk}, \textbf{LINE}~\cite{tang2015line}, \textbf{Node2vec}~\cite{grover2016node2vec}, \textbf{Metapath2vec++}~\cite{dong2017metapath2vec}, and \textbf{HEER}~\cite{shi2018easing}.  We applied this group of baselines on constructed Chinese character variation graph to get graph based character embeddings. Specifically, \textbf{Metapath2vec++} required a human-defined metapath scheme to guide the  random walks. We tried 4 different metapaths for this experiment:(1) \textbf{M2V}$\mathbf{_{P}}$ (only walking on pinyin (phonetic) relations); (2) \textbf{M2V}$\mathbf{_{S}}$ (only walking on stroke (glyph) relations); (3) \textbf{M2V}$\mathbf{_{Z}}$ (only walking on Zhengma (glyph) relations); (4) \textbf{M2V}$\mathbf{_{C}}$ (alternately walking on glyph and phonetic relations). We reported the best results from these four metapaths, denoted as \textbf{M2V}$\mathbf{_{Max}}$.

\textit{Spelling Correction Baseline}: \textbf{Pycorrector}\footnote{https://github.com/shibing624/pycorrector} based on n-gram language model~\cite{yu2014chinese}.

\textit{Comparison Group}: we compared the performances of several variants of the proposed method in order to highlight our technical contributions. There were 3 comparison groups conducted. \textbf{\model}$\mathbf{_{Graph}}$: we only used {\emb} graph embedding. \textbf{\model}$\mathbf{_{Naive}}$: we simply concatenated {\emb} graph embedding and skip-gram textual embedding (a naive version).
\textbf{\model}$\mathbf{_{Original}}$: the proposed {\model} model. 

For a fair comparison, the dimension\footnote{The initial dimension of {\model}$\mathbf{_{Naive}}$ and {\model}$\mathbf{_{Original}}$ is 256, so we used a fully connected layer to reduce its dimension to 128.} of all embedding models was 128. A single layer of CNN classification model\footnote{The filter sizes of CNN is $3, 4, 5$, and the filter number is $128$, dropout ratio is $0.1$.} was used for spam detection task.

\begin{figure*}[!htbp]\centering
 	\includegraphics[width=2.0\columnwidth]{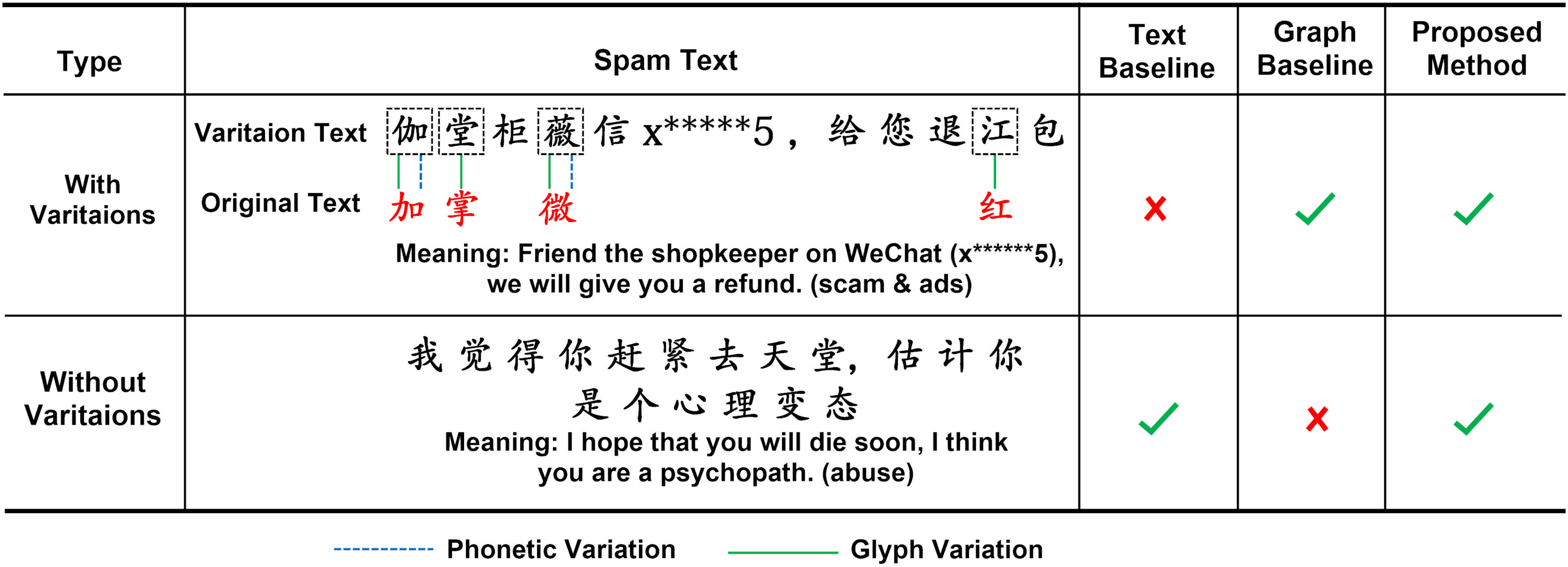}
 	\caption{Two typical examples for {\problem} task}
 	\label{fig:case2}
\end{figure*}

\subsection{Experiment Result and Analysis}
The text spam detection task performances of different models were reported in Table \ref{tab:result}. Based on the experiment results, we had the following observations: 

\textbf{\model~vs. Baselines}. (1) {\model}$\mathbf{_{Original}}$ outperformed the baseline models for all evaluation metrics on both datasets, which indicated the proposed {\model} model can effectively address the {\problem} problem. (2) On review dataset, the leading gap between {\model}$\mathbf{_{Original}}$ and other baselines was greater. A possible explanation was that, the review spam text usually had richer content and more complex variation patterns than SMS spam text. Therefore, a good variation representation model may have certain advantages.

\textbf{Chinese vs. General}. (1) Compared to classical textual embedding models (Skipgram and GloVe), the Chinese embedding models showed their advantages, especially on review dataset. This result indicated that the characteristic knowledge of Chinese can help to detect spam text. (2) ELMo was able to learn both the semantic and contextualized information, and it achieved a good performance in text baseline group. 

\textbf{Graph vs. Text}. Generally, the graph based baselines outperformed the textual based baselines (including general and Chinese). This observation indicated: (1) the variation knowledge of Chinese character can be critical for {\problem} problem. (2) The proposed character variation graph can provide critical information for Chinese character representation learning. (3) Compared to other graph based baselines, {\model}$\mathbf{_{Graph}}$ was superior, which proved the effectiveness of {\emb} algorithm, and the proposed variation family can characterize and predict useful variation patterns for {\problem} problem.

\textbf{Chinese Character Encodings}. (1) In Chinese textual embedding baseline group, JWE (radical based) and Cw2vec (stroke based) didn't perform well, which indicated employing a single kind of glyph-based information can be insufficient for Chinese variation representation learning. Similarly, in graph based baseline group, the performances of {M2V}$\mathbf{_{P}}$, {M2V}$\mathbf{_{S}}$ and {M2V}$\mathbf{_{Z}}$ (employed only one encoding relation on the constructed graph) were still unsatisfactory. The results revealed that an individual encoding method cannot comprehensively encode a character, we should consider various kinds of variation information simultaneously. (2) The performance of {M2V}$\mathbf{_{C}}$ (integrated all relations based on a predefined metapath pattern) was still inferior. This result indicated a human-defined rule cannot effectively integrate all relationships in a complex graph.

\textbf{Representation vs. Spelling Correction}. Pycorrector performed poorly in experiment, and other baselines outperformed this approach, which proved the spelling correction method is not capable for {\problem} problem.

\textbf{Variants of \model~model}. For variants of the proposed method, the results showed that (1) by combining the semantic and sequential information, the task performances can improve; (2) simply concatenating graph and text embeddings cannot generate a satisfactory joint representation. (3) The proposed {\model} model can successfully capture the variation, semantic, and sequential information for character representation learning.

\subsection{Case Study}
To gain an insightful understanding regarding the variation representation of the proposed method, we conduct qualitative analysis by performing the case studies of character similarities. As shown in Table \ref{tab:case}, for exemplary characters, the most similar characters, based on skipgram embedding (general textual based baseline), are all semantically similar or/and context-related. Meanwhile, based on Cw2vec embedding (most recent Chinese embedding baseline), all similar characters for target characters are also semantically similar or/and context-related. Unsurprisingly, for each target character, all similar characters based on {\emb} model (best performed graph embedding model), are glyph and phonetic similar characters. The proposed {\model} model can achieve a comprehensive coverage from variation, semantic and context viewpoints. For instance, in its top 3 similar characters for ``运(move)'',  ``转(transmit)'' is a semantic and context similar character, and ``云(cloud)'' is a glyph and phonetic similar character. Furthermore, {\model} model can capture complicated compound similarity between Chinese characters, for instance, ``悚(afraid)'' is a glyph, semantic, and context similar character for ``惊(shock)''.  This also explains why {\model} model performs well to address the {\problem} problem.

Figure \ref{fig:case2} depicts two typical examples in the experimental datasets. For the spam text with variations, spammers used character variations to create camouflaged expressions. For instance, using glyph variation ``江(river)'' to replace ``红(red)'', and glyph-phonetic compound variation ``薇(osmund)'' to replace ``微(micro)''. The classical text embedding models may fail to identify this kind of spam texts. With the mining of character variation graph, the graph based approaches can be successful to capture these changes. For spam text without variations, classification models need more semantic and contextual information, and the text based methods can be suitable for this kind of spam texts. The proposed {\model} model is able to detect both two kinds of spam texts effectively, and experiment results proved  {\model} can successfully model Chinese variational, semantic and contextualized representations for {\problem} task.
\section{Conclusion}
In this paper, we propose a StoneSkipping model for Chinese spam detection. The performance of the proposed method is comprehensively evaluated in two real world datasets with challenging experimental setting. The results of experiments show that the proposed model significantly outperforms a number of state-of-the-art methods. Meanwhile, the case study empirically proves that the proposed model can successfully capture the Chinese variation, semantic, and contextualized information, which can be essential for {\problem} problem. In the future, we will investigate more sophisticated methods to improve {\model}'s performance, e.g., enable self-attention mechanism for contextualized information modelling. 

\section*{Acknowledgments}
This work is supported by the National Natural Science Foundation of China (61876003, 81971691), the China Department of Science and Technology Key Grant (2018YFC1704206), and Fundamental Research Funds for the Central Universities (18lgpy62).

\bibliography{emnlp-ijcnlp-2019}
\bibliographystyle{acl_natbib}
\end{CJK}
\end{document}